\documentclass[runningheads]{llncs}
\usepackage{graphicx}
\usepackage{amsmath,amssymb,amsfonts}
\usepackage{algorithmic}
\usepackage{hyperref}
\usepackage{multirow}
\usepackage{rotating}
\usepackage{subcaption}
\usepackage{tabularx,ragged2e}
\hypersetup{
    colorlinks = true,
}
\begin{document}
%

\title{MedCLIP-SAM: Bridging Text and Image Towards Universal Medical Image Segmentation}
%
\titlerunning{MedCLIP-SAM for Universal Medical Image Segmentation}

%
\author{Taha Koleilat\inst{1} \and
Hojat Asgariandehkordi\inst{1}\and
Hassan Rivaz\inst{1}\and
Yiming Xiao\inst{2}}

%
\authorrunning{T. Koleilat et al.}
%
\institute{Department of Electrical and Computer Engineering, Concordia University, Montreal, Canada \\
\email{\{taha.koleilat,hojat.asgariandehkordi,hassan.rivaz\}@concordia.ca} 
\and
Department of Computer Science and Software Engineering, Concordia University, Montreal, Canada
\\
\email{yiming.xiao@concordia.ca}}


%
\maketitle              
\begin{abstract}
Medical image segmentation of anatomical structures and pathology is crucial in modern clinical diagnosis, disease study, and treatment planning. To date, great progress has been made in deep learning-based segmentation techniques, but most methods still lack data efficiency, generalizability, and interactability. Consequently, the development of new, precise segmentation methods that demand fewer labeled datasets is of utmost importance in medical image analysis. Recently, the emergence of foundation models, such as CLIP and Segment-Anything-Model (SAM), with comprehensive cross-domain representation opened the door for interactive and universal image segmentation. However, exploration of these models for data-efficient medical image segmentation is still limited, but is highly necessary. In this paper, we propose a novel framework, called MedCLIP-SAM that combines CLIP and SAM models to generate segmentation of clinical scans using text prompts in both zero-shot and weakly supervised settings. To achieve this, we employed a new Decoupled Hard Negative Noise Contrastive Estimation (DHN-NCE) loss to fine-tune the BiomedCLIP model and the recent gScoreCAM to generate prompts to obtain segmentation masks from SAM in a zero-shot setting. Additionally, we explored the use of zero-shot segmentation labels in a weakly supervised paradigm to improve the segmentation quality further. By extensively testing three diverse segmentation tasks and medical image modalities (breast tumor ultrasound, brain tumor MRI, and lung X-ray), our proposed framework has demonstrated excellent accuracy. Code is available at \url{https://github.com/HealthX-Lab/MedCLIP-SAM}.

\end{abstract}
\section{Introduction}

With the increasing availability of radiological technologies, there is a pressing need for accurate and efficient medical image segmentation to aid the study, diagnosis, and treatment of various medical conditions \cite{siuly2016medical}. Deep learning (DL) techniques have been established as state-of-the-art in the domain, but current methods often face three major limitations, hindering their widespread clinical adoption. First, the lack of large well-annotated data sets is a major bottleneck for DL model development. Second, the lack of 
interactability and interpretability limits the credence of the methods. Lastly, most trained models are task- and contrast/modality-specific with low flexibility. While many self- and weakly supervised methods \cite{baevski2023efficient,chen2020big,taleb2021multimodal} have been proposed to tackle training data efficiency and explainable AI (XAI) methods (e.g., uncertainty estimation \cite{loquercio2020general,liu2020simple} and saliency map \cite{arun2021assessing,bae2020rethinking}) are being actively investigated, cross-domain generalization has been a challenge. Recently, the introduction of foundation models, such as the CLIP (Contrastive Language-Image Pre-Training) \cite{radford2021learning} and SAM (Segment Anything Model) \cite{kirillov2023segment} opened the door for interactive and universal medical image segmentation. To date, several groups have endeavored to adapt CLIP and SAM for radiological tasks from natural images, notably the development of BiomedCLIP \cite{zhang2023largescale} and MedSAM \cite{Ma2023SegmentAI}, which were pre-trained on millions of biomedical data. However, more efficient parameter fine-tuning methods can be beneficial to further boost the performance of these foundation models in radiological applications. On the other hand, with a strong interest in SAM, which requires interactive prompts to guide segmentation, a few techniques were proposed to fine-tune SAM without prompts \cite{chen2024unsam,hu2023efficiently}, generate prompts through Class Activation Map (CAM) from classification tasks \cite{li2024clipsam,li2023clip,liu2024weakly}, and to refine its output based on weak supervision \cite{yang2023foundation,chen2023segment,huang2023push}. Still at the nascent phase, using foundation models for interactive and universal medical image segmentation necessitates additional investigation and is of significant interest.

To address the aforementioned needs, we present MedCLIP-SAM, a novel framework that leverages BiomedCLIP \cite{zhang2023largescale} and SAM \cite{kirillov2023segment} for text-prompt-based interactive and universal medical image segmentation in both zero-shot and weakly supervision settings. The contributions of this work are threefold: \textbf{First}, we proposed a novel CLIP training/fine-tuning method, called the Decoupled Hard Negative Noise Contrastive Estimation (DHN-NCE). \textbf{Second}, we proposed a zero-shot medical segmentation method by combining CLIP and SAM in radiological tasks for the first time. \textbf{Lastly}, a weakly-supervised strategy was explored with the attempt to further refine zero-shot segmentation results, and the full proposed technique was extensively validated on three different segmentation tasks and modalities (breast tumor segmentation in ultrasound, brain tumor segmentation in MRI, and lung segmentation in chest X-ray).

\section{Methods and Materials}
\label{section:methods}

\begin{figure}[h]
\includegraphics[width=\textwidth]{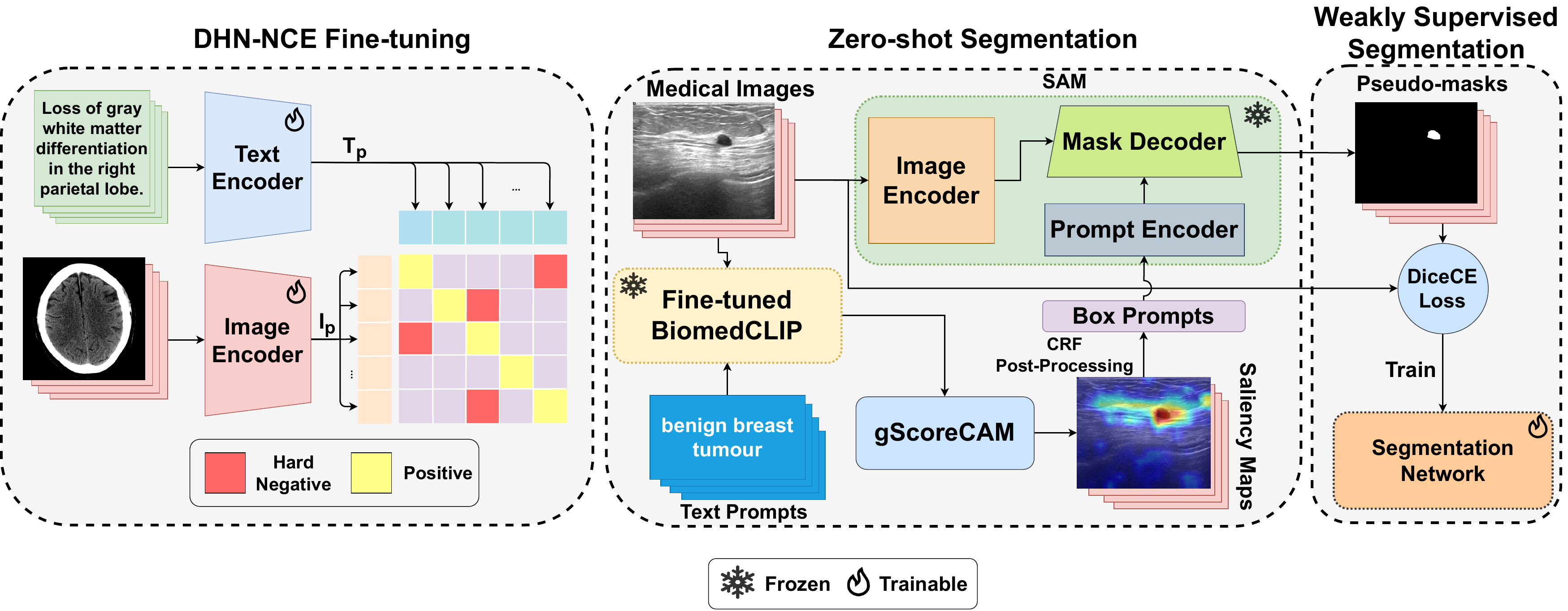}
\caption{An overview of the proposed MedCLIP-SAM framework.} 
\label{fig:System Figure}
\end{figure}

An overview of the proposed MedCLIP-SAM framework is presented in Fig. \ref{fig:System Figure}, organized into three distinct stages:  BiomedCLIP fine-tuning employing our new DHN-NCE loss, zero-shot segmentation guided by text-prompts, and weakly supervised segmentation for potential label refinement.

\subsection{Efficient BiomedCLIP Fine-Tuning with the DHN-NCE loss}
\subsubsection{Decoupled Hard Negative Noise Contrastive Estimation loss} 
A CLIP model is trained in large datasets of images and the corresponding texts. Specifically, an image encoder and a text encoder are used to extract features of images and texts and project them into vectors of the same dimension, $\mathbf{I}_{p,i}$ and $\mathbf{T}_{p,i}$, respectively. Then, through contrastive learning, an embedding space shared by the image and text vectors is learned so that similar pairs (an image and its description) are closer together and dissimilar ones are farther apart. While BiomedCLIP \cite{zhang2023largescale} was trained on medical charts/images and clinical texts, further fine-tuning can effectively benefit medical image-specific tasks. In CLIP training with the conventional InfoNCE loss \cite{oord2018representation}, the \textit{negative-positive-coupling (NPC)} effect \cite{yeh2022decoupled} can lead to sub-optimal learning efficiency, particularly in small batch sizes while for medical images, more nuanced discrimination between cases within the same imaging categories can be difficult. To solve these, we propose the Decoupled Hard Negative Noise Contrastive Estimation (DHN-NCE) loss, which 1) combines the InfoNCE loss \cite{oord2018representation} with hard negative sampling \cite{robinson2021contrastive} to focus on ``close samples'' and 2) adds decoupling contrastive learning \cite{yeh2022decoupled} by removing the positive term in the denominator to allow smaller batch sizes. Specifically, the loss function $\mathcal{L}_{DHN-NCE}$ uses weighting functions ($\mathcal{W}_{\mathbf{I}_{p,i}\mathbf{T}_{p,j}}^{v \rightarrow t}$,$\mathcal{W}_{\mathbf{T}_{p,i}\mathbf{I}_{p,j}}^{t \rightarrow v}$) to increase the penalty for negatives that happen to be very close to the anchor through image-to-text and text-to-image hardness parameters $\beta_1,\beta_2 \geq 0$. Here, $t \rightarrow v$ means text-to-image, and   $v \rightarrow t$ denotes image-to-text.

\begin{equation}
    \mathcal{L}^{v \rightarrow t} = -\sum\limits_{i=1}^B\frac{\mathbf{I}_{p,i}\mathbf{T}_{p,i}^\top}{\tau} +  \sum\limits_{i=1}^Blog\left(\sum\limits_{j \neq i}e^{\mathbf{I}_{p,i}\mathbf{T}_{p,j}^\top/\tau}\mathcal{W}_{\mathbf{I}_{p,i}\mathbf{T}_{p,j}}^{v \rightarrow t}\right)
\end{equation}
\begin{equation}
    \mathcal{L}^{t \rightarrow v} = -\sum\limits_{i=1}^B\frac{\mathbf{T}_{p,i}\mathbf{I}_{p,i}^\top}{\tau} + \sum\limits_{i=1}^Blog\left(\sum\limits_{j \neq i}e^{\mathbf{T}_{p,i}\mathbf{I}_{p,j}^\top/\tau}\mathcal{W}_{\mathbf{T}_{p,i}\mathbf{I}_{p,j}}^{t \rightarrow v}\right)
\end{equation}
\begin{equation}
    \mathcal{L}_{DHN-NCE} = \mathcal{L}^{v \rightarrow t} + \mathcal{L}^{t \rightarrow v} 
\end{equation}
where $B$ is the batch size, $\tau$ is the temperature parameter, and the hardness weighting formulas are as follows:
\begin{equation}
    \mathcal{W}_{\mathbf{I}_{p,i}\mathbf{T}_{p,j}}^{v \rightarrow t} = (B-1)\times\frac{e^{\beta_1\mathbf{I}_{p,i}\mathbf{T}_{p,j}/\tau}}{\sum_{k \neq i}e^{\beta_1\mathbf{I}_{p,i}\mathbf{T}_{p,k}/\tau}}
\end{equation}
\begin{equation}
    \mathcal{W}_{\mathbf{T}_{p,i}\mathbf{I}_{p,j}}^{t \rightarrow v} = (B-1)\times\frac{e^{\beta_2\mathbf{T}_{p,i}\mathbf{I}_{p,j}/\tau}}{\sum_{k \neq i}e^{\beta_2\mathbf{T}_{p,i}\mathbf{I}_{p,k}/\tau}}
\end{equation}

\subsubsection{BiomedCLIP fine-tuning} 
We utilized the public MedPix dataset with different radiological modalities to fine-tune the BiomedCLIP model \cite{zhang2023largescale} with DHN-NCE loss. Here, we used the base Vision Transformer and PubMedBERT \cite{zhang2023largescale} as the image and text encoders. We cleaned the MedPix dataset by stripping off any special characters, leading and trailing white spaces, and deleting samples with captions of less than 20 characters. All images were resized to 224 $\times$ 224 pixels and normalized by the RGB channel means and standard deviations used in the original CLIP model \cite{radford2021learning}. After performing an 85\%:15\% split, we ended up with 20,292 training images and 3,515 images for validation. Here, we chose a low learning rate of 1E-6 with a decay rate of 50\%, and fine-tuning was done on batches of 64 samples.

\subsection{Zero-shot and Weakly Supervised Medical Image Segmentation}
With a fine-tuned BiomedCLIP model, we proposed a zero-shot universal medical image segmentation strategy, which leverages the recent XAI technique, gScoreCAM \cite{chen2022gScoreCAM} that provides visual saliency maps of text prompts in corresponding images for CLIP models. While gScoreCAM was shown to outperform gradCAM in natural images in accuracy and specificity, we adopted it in radiological tasks for the first time. Here, for an input image and a text prompt for the target anatomy/pathology, we first obtained an initial, coarse segmentation by post-processing the gScoreCAM map with a conditional random field (CRF) filter \cite{pmlr-v28-kraehenbuehl13}, which was then used to obtain a bounding box for SAM to produce a pseudo-mask as zero-shot segmentation. In the attempt to further enhance the accuracy of zero-shot segmentation, we used the resulting pseudo-masks to train a Residual UNet \cite{Zhang_2018} in a weakly supervised setting.

\subsection{Datasets, Experimental Setup, and Validation Metrics}
\subsubsection{BiomedCLIP fine-tuning performance}
We validated the quality of BiomedCLIP fine-tuning by the accuracy of top 1 and top 2 matching retrievals for both image-to-text and text-to-image directions in the ROCO (Radiology Objects in COntext) dataset \cite{Pelka2018RadiologyOI} which contains $\approx$ 7,042 multi-modal medical images spanning a myriad of clinical cases. We executed the experiments for 5 runs with a batch size of 50 with shuffling to ensure random bagging of different texts and images within a batch (thus, we get 70,420 shuffled examples). We compared different loss functions for fine-tuning, including the InfoNCE loss \cite{oord2018representation}, DCL \cite{yeh2022decoupled}, HN-NCE \cite{rdk+23}, and our DHN-NCE loss. For a fair comparison, we trained all the strategies using the same hyperparameters ($\tau$ = 0.6, learning rate = 1E-6). For HN-NCE and DHN-NCE, we use the same hardness  $\beta_1$ = $\beta_2$ = 0.15. As baselines, we also included the results of pre-trained BiomedCLIP \cite{zhang2023largescale}, PMC-CLIP \cite{lin2023pmcclip}, and CLIP \cite{radford2021learning}.

\subsubsection{Image segmentation accuracy}
To validate the zero-shot and weakly supervised segmentation results, as well as different design components of the MedCLIP-SAM framework, we used three public datasets (three different modalities) with segmentation ground truths (segmentation of breast tumor, brain tumor, and lung), which were split for training, validation, and testing. These datasets with their divisions include:
\begin{itemize}
    \item \textbf{Breast Tumor Ultrasound}: Breast Ultrasound Images dataset (BUSI) \cite{ALDHABYANI2020104863} with 600 benign and malignant tumors images for training only; 65 and 98 images from the UDIAT\cite{Byra2020-gg} dataset for validation and testing, respectively.
    \item \textbf{Brain Tumor MRI}: Brain Tumor dataset from \cite{Cheng2017} consisting of 1,462, 400, and 400 T1-weighted MRIs for training, validation and testing respectively. 
    \item \textbf{Lung Chest X-ray}: COVID-19 Radiography Database (COVID-QU-Ex) \cite{9144185,RAHMAN2021104319} with  16,280, 1,372, and 957 Chest X-ray scans (normal, lung opacity, viral pneumonia, and COVID-19 cases) for training, validation, and testing.
\end{itemize}

\noindent With these datasets, we conducted a detailed comparison of the segmentation quality for the initial labels based on CRF-processed gScoreCAM results, zero-shot pseudo-masks, and weakly supervised results on the aforementioned testing sets. As ablation studies for zero-shot segmentation, we investigated \textbf{1)} the impacts of BiomedCLIP fine-tuning and \textbf{2)} the choice of gScoreCAM vs. gradCAM. The ablation studies were performed on the test set of each of the three aforementioned datasets. For a fair comparison, we utilized the same SAM model, target layer, text prompts, and CAM settings of the top 60 channels for all data across different variations. In all experiments, Intersection over Union (IoU), Dice Similarity Coefficient (DSC), and area under the ROC curve (AUC) were used, and paired-sample t-tests were performed to confirm the observations and trends. Here, a p-value $<$ 0.05 indicates a statistically significant difference.

\section{Results}

\subsection{Cross-modal retrieval accuracy and gScoreCAM vs. gradCAM:}
 The accuracy of cross-modal retrieval (text-to-image and image-to-text) for the ROCO dataset \cite{Pelka2018RadiologyOI} is shown in Table \ref{tab:zero acc} across different losses for fine-tuning BiomedCLIP, with three pre-trained CLIP models as baselines. Paired McNemar statistical tests show that our DHN-NCE significantly outperformed other existing loss functions and pre-trained baseline models ($p<$0.01). In Table \ref{tab:ablation}, we present the accuracy evaluation for our MedCLIP-SAM zero-shot segmentation with different setups (Pre-trained BiomedCLIP vs. fine-tuned BiomedCLIP and gScoreCAM vs. GradCAM). The comparison demonstrated the great advantages of using gScoreCAM over GradCAM to generate bounding-box prompts for SAM ($p<$1E-4). Additionally, the benefit of fine-tuning BiomedCLIP with our DHN-NCE loss is further validated with improved segmentation quality across different tasks and image modalities ($p<0.05$). 

 \begin{table}[h]
\centering
\scriptsize
\caption{Top-K cross-modal retrieval accuracy (mean$\pm$std) for CLIP models.}
\begin{tabular}{|c|c|cc|cc|}
\hline
\multirow{2}{*}{Model} & \multirow{2}{*}{Version} & \multicolumn{2}{c|}{$image \rightarrow text 
 $ (\%)}                         & \multicolumn{2}{c|}{$text \rightarrow image 
 $ (\%)}                         \\ \cline{3-6} 
                       &                          & \multicolumn{1}{c|}{Top-1}          & Top-2          & \multicolumn{1}{c|}{Top-1}          & Top-2          \\ \hline
\multirow{5}{*}{BiomedCLIP \cite{zhang2023largescale}} &
  Pre-trained &
  \multicolumn{1}{c|}{81.83 $\pm$ 0.20} &
  92.79 $\pm$ 0.13 &
  \multicolumn{1}{c|}{81.36 $\pm$ 0.48} &
  92.27 $\pm$ 0.14 \\
                       & InfoNCE \cite{oord2018representation}                 & \multicolumn{1}{c|}{84.21 $\pm$ 0.35} & 94.47 $\pm$ 0.19 & \multicolumn{1}{c|}{85.73 $\pm$ 0.19} & 94.99 $\pm$ 0.16 \\
                       & DCL \cite{yeh2022decoupled}                   & \multicolumn{1}{c|}{84.44 $\pm$ 0.37} & 94.68 $\pm$ 0.19 & \multicolumn{1}{c|}{85.89 $\pm$ 0.16} & 95.09 $\pm$ 0.19 \\
                       & HN-NCE \cite{rdk+23}                 & \multicolumn{1}{c|}{84.33 $\pm$ 0.35} & 94.60 $\pm$ 0.19 & \multicolumn{1}{c|}{85.80 $\pm$ 0.17} & 95.10 $\pm$ 0.19 \\
 &
  \textbf{DHN-NCE (Ours)} &
  \multicolumn{1}{c|}{\textbf{84.70 $\pm$ 0.33}} &
  \textbf{94.73 $\pm$ 0.16} &
  \multicolumn{1}{c|}{\textbf{85.99 $\pm$ 0.19}} &
  \textbf{95.17 $\pm$ 0.19} \\ \hline
CLIP \cite{radford2021learning}                   & Pre-trained              & \multicolumn{1}{c|}{26.68 $\pm$ 0.30} & 41.80 $\pm$ 0.19 & \multicolumn{1}{c|}{26.17 $\pm$ 0.20} & 41.13 $\pm$ 0.20 \\ \hline
PMC-CLIP \cite{lin2023pmcclip}            & Pre-trained              & \multicolumn{1}{c|}{75.47 $\pm$ 0.37} & 87.46 $\pm$ 0.11 & \multicolumn{1}{c|}{76.78 $\pm$ 0.11} & 88.35 $\pm$ 0.19 \\ \hline
\end{tabular}
\newline
\label{tab:zero acc}
\end{table}
\subsection{Zero-shot and Weakly Supervised Segmentation}

\begin{figure*}[h]
    \centering
    \begin{subfigure}[b]{0.19\textwidth}
        \centering
        \includegraphics[width=\textwidth]{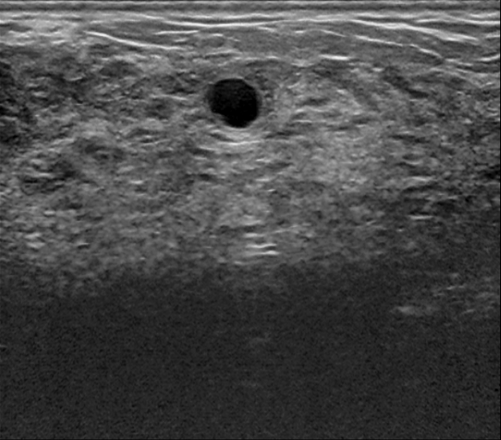}
    \end{subfigure}
    \begin{subfigure}[b]{0.19\textwidth}  
        \centering 
        \includegraphics[width=\textwidth]{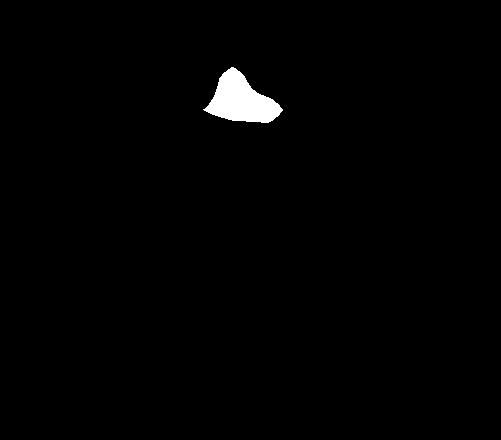}
    \end{subfigure}
    \begin{subfigure}[b]{0.19\textwidth}   
        \centering 
        \includegraphics[width=\textwidth]{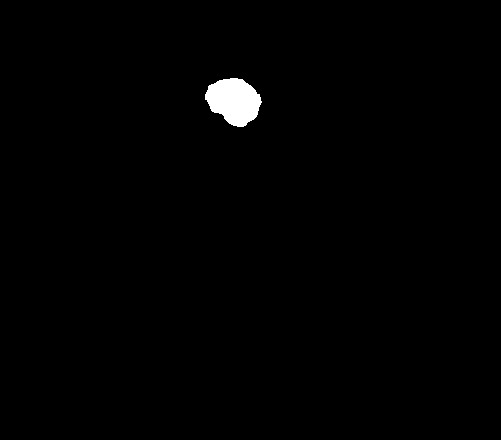}
    \end{subfigure}
    \begin{subfigure}[b]{0.19\textwidth}   
        \centering 
        \includegraphics[width=\textwidth]{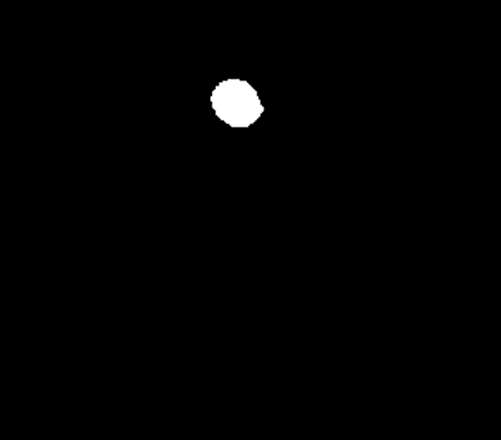}
    \end{subfigure}
    \begin{subfigure}[b]{0.19\textwidth}   
        \centering 
        \includegraphics[width=\textwidth]{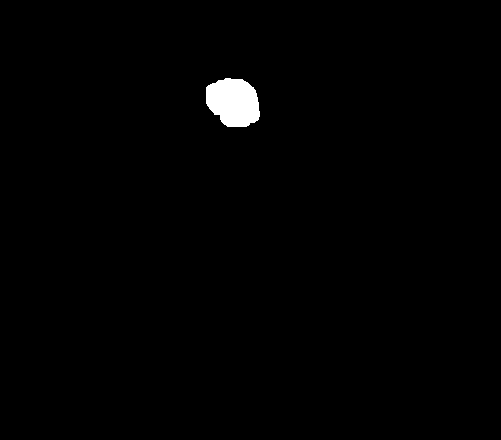}
    \end{subfigure}
    
    
    \begin{subfigure}[b]{0.19\textwidth}
        \centering
        \includegraphics[width=\textwidth]{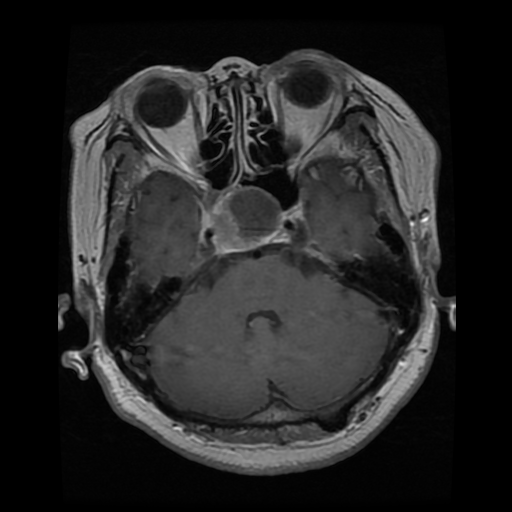}
    \end{subfigure}
    \begin{subfigure}[b]{0.19\textwidth}   
        \centering 
        \includegraphics[width=\textwidth]{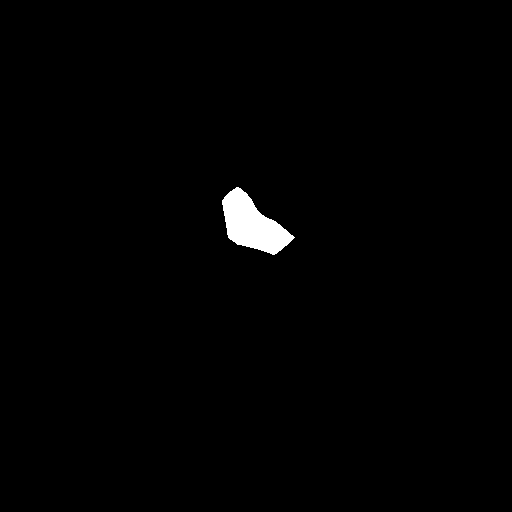}
    \end{subfigure}
    \begin{subfigure}[b]{0.19\textwidth}   
        \centering 
        \includegraphics[width=\textwidth]{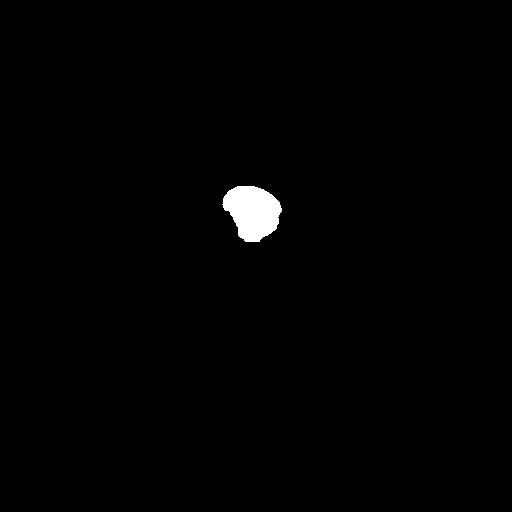}
    \end{subfigure}
    \begin{subfigure}[b]{0.19\textwidth}  
        \centering 
        \includegraphics[width=\textwidth]{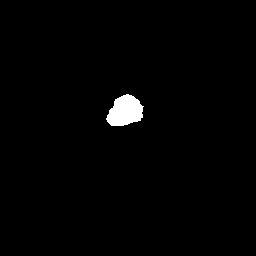}
    \end{subfigure}
    \begin{subfigure}[b]{0.19\textwidth}   
        \centering 
        \includegraphics[width=\textwidth]{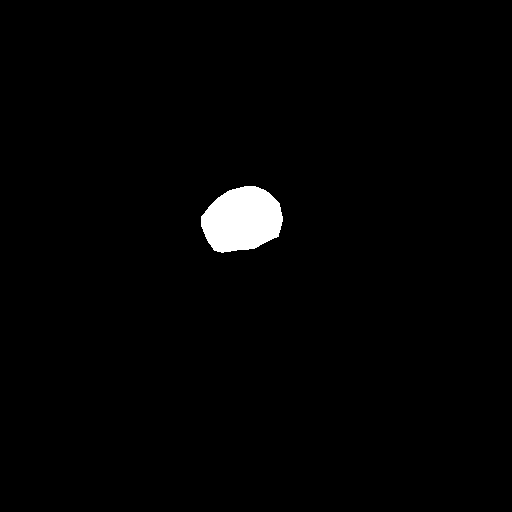}
    \end{subfigure}
    
    
    \begin{subfigure}[b]{0.19\textwidth}
        \centering
        \includegraphics[width=\textwidth]{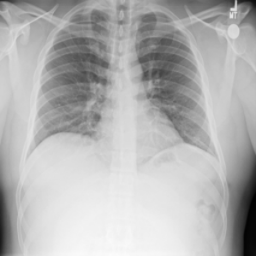}
        \caption{Image}
    \end{subfigure}
    \begin{subfigure}[b]{0.19\textwidth}   
        \centering 
        \includegraphics[width=\textwidth]{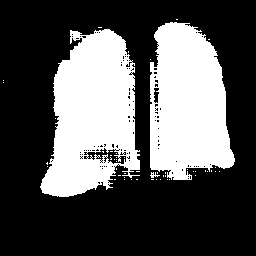}
        \caption{CLIP-CRF}
    \end{subfigure}
    \begin{subfigure}[b]{0.19\textwidth}   
        \centering 
        \includegraphics[width=\textwidth]{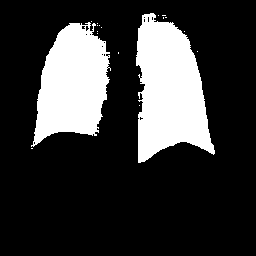}
        \caption{Zero-Shot}
    \end{subfigure}
    \begin{subfigure}[b]{0.19\textwidth}   
        \centering 
        \includegraphics[width=\textwidth]{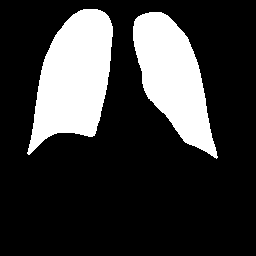}
        \caption{WSS}
    \end{subfigure}
    \begin{subfigure}[b]{0.19\textwidth}  
        \centering 
        \includegraphics[width=\textwidth]{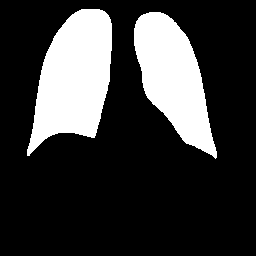}
        \caption{GT}
    \end{subfigure}
    
    \caption{Qualitative comparison of segmentation results. CLIP-CRF=CRF processed BiomedCLIP saliency map and WSS=weakly supervised segmentation.}
    \label{fig:seg-illustration}
\end{figure*}

In Table \ref{tab:full-seg}, we present segmentation accuracy for our proposed method in zero-shot and weakly supervised settings, with fully supervised segmentation as a reference. Note that for zero-shot results, we include a comparison between initial labels generated by gScoreCAM-based saliency maps (``Saliency Maps") and pseudo-masks from SAM (``Saliency Maps + SAM"). Combining BiomedCLIP and SAM demonstrates clear advantages, notably improving segmentation quality for all metrics ($p < 0.05$). Comparing zero-shot results to weakly supervised segmentation, we observe general improvements for X-ray-based lung segmentation. However, the impact on tumor segmentation in breast ultrasound and brain MRI remains unclear, with an AUC boost of $\sim$2\% only for breast ultrasound. While fully supervised DL models currently provide state-of-the-art accuracy for medical image segmentation, our MedCLIP-SAM zero-shot segmentation outperformed ResUNet-based full supervision for breast ultrasound and brain MRI segmentation. Lung X-ray segmentation, however, showed superior accuracy with the fully supervised method across all metrics. Finally, to provide a qualitative assessment, exemplary segmentation results for zero-shot and weakly supervised settings are shown in Fig. \ref{fig:seg-illustration} against the original image and ground truths (GTs) across all segmentation tasks.

\begin{table}[h]
\scriptsize
\centering
\caption{Comparison of zero-shot segmentation accuracy (mean$\pm$std) with SAM based on the pre-trained and fine-tuned BiomedCLIP models using gScoreCAM vs. GradCAM techniques for bounding-box generation.}
\begin{tabular}{|c|c|c|c|c|c|}
\hline
Modality & Model                 & CAM       & IoU (\%)      & DSC (\%)       & AUC (\%)       \\ \hline
\multirow{4}{*}{Breast Ultrasound}         & \multirow{2}{*}{BiomedCLIP} & gScoreCAM & 56.24 $\pm$ 9.25 & 66.03 $\pm$ 8.77 & 78.59 $\pm$ 6.38 \\
        &                       & GradCAM   & 18.16 $\pm$ 9.67         & 23.99 $\pm$ 8.24         & 60.12 $\pm$ 6.36          \\ \cline{2-3}
        & \multirow{2}{*}{\textbf{Ours}} & \textbf{gScoreCAM} & \textbf{57.97 $\pm$ 8.59} & \textbf{67.82 $\pm$ 8.26} & \textbf{79.31 $\pm$ 6.84} \\
        &                       & GradCAM   & 20.79 $\pm$ 9.32          &   25.65 $\pm$ 7.81        & 62.54 $\pm$ 5.22         \\ \hline
\multirow{4}{*}{Brain MRI} & \multirow{2}{*}{BiomedCLIP} & gScoreCAM & 48.87 $\pm$ 6.71 & 65.13 $\pm$ 5.98  & 79.69 $\pm$ 6.12 \\
        &                       & GradCAM   & 26.69 $\pm$ 7.45          & 32.03  $\pm$ 5.23        & 76.04  $\pm$ 7.86        \\ \cline{2-3}
        & \multirow{2}{*}{\textbf{Ours}} & \textbf{gScoreCAM} & \textbf{50.30 $\pm$ 5.94}  & \textbf{66.72 $\pm$ 5.27} & \textbf{81.35 $\pm$ 6.33} \\
        &                       & GradCAM   & 27.07 $\pm$ 7.29           & 33.10 $\pm$ 6.91         & 78.72  $\pm$ 7.16        \\ \hline
\multirow{4}{*}{Lung X-ray}  & \multirow{2}{*}{BiomedCLIP} & gScoreCAM & 47.95 $\pm$ 10.37 & 63.21 $\pm$ 11.70  & 77.53 $\pm$ 5.49 \\
        &                       & GradCAM   & 22.79 $\pm$ 7.35 & 35.21 $\pm$ 10.75  & 60.19 $\pm$ 4.73           \\ \cline{2-3}
        & \multirow{2}{*}{\textbf{Ours}} & \textbf{gScoreCAM} & \textbf{49.06 $\pm$ 9.22} & \textbf{64.49 $\pm$ 9.09} & \textbf{78.54 $\pm$ 5.64} \\
        &                       & GradCAM   & 26.45 $\pm$ 8.39          & 39.75 $\pm$ 8.44          & 62.95 $\pm$ 5.71          \\ \hline
\end{tabular}
\label{tab:ablation}
\end{table}

\begin{table}[h]
\scriptsize
\centering
\caption{Segmentation accuracy (mean$\pm$std) for zero-shot and weakly supervised methods against a fully supervised baseline.}
\begin{tabular}{|c|c|c|c|c|c|}
\hline
Modality & Model                       & IoU (\%)     & DSC (\%)     & AUC (\%)     \\ \hline
\multirow{4}{*}{Breast Ultrasound}        & Saliency Maps& 40.43 $\pm$ 8.34 & 51.82 $\pm$ 9.60 & 73.77 $\pm$ 7.54  \\ 
        &Saliency Maps + SAM &  \textbf{57.97} $\pm$ 8.59 & \textbf{67.82 $\pm$ 8.26} & 79.31 $\pm$ 6.84 \\ 
         &           Weak supervision-ResUNet \cite{Zhang_2018}& 41.68 $\pm$ 5.63             &  58.62 $\pm$ 5.66            &  81.44 $\pm$ 4.22             \\ 
                    & Full supervision-ResUNet \cite{Zhang_2018} &  53.15 $\pm$ 8.36               & 67.29 $\pm$ 7.84             & \textbf{84.74 $\pm$ 5.09}             \\
                    \hline
\multirow{4}{*}{Brain MRI} & Saliency Maps& 39.12 $\pm$ 6.11  & 53.06 $\pm$ 6.34 & 75.89 $\pm$ 6.92 \\ 
        &Saliency Maps + SAM & \textbf{50.30 $\pm$ 5.94}  & \textbf{66.72 $\pm$ 5.27} & \textbf{81.35 $\pm$ 6.33} \\
        &                    Weak supervision-ResUNet \cite{Zhang_2018}&  42.17 $\pm$ 8.67 &  58.80 $\pm$ 8.63                 & 78.25 $\pm$ 5.32             \\ & Full supervision-ResUNet \cite{Zhang_2018} &   45.93 $\pm$ 7.68   & 62.57 $\pm$ 7.20                       &  79.85 $\pm$ 4.87            \\\hline
\multirow{4}{*}{Lung X-ray}  & Saliency Maps &  35.04 $\pm$ 8.40  & 49.54 $\pm$ 9.18 & 71.94 $\pm$ 6.21 \\ 
        & Saliency Maps + SAM & 49.06 $\pm$ 9.22 & 64.49 $\pm$ 9.09 & 78.54 $\pm$ 5.64 \\  
        &                   Weak supervision-ResUNet \cite{Zhang_2018}&  76.46 $\pm$ 12.03            &  86.07 $\pm$ 8.61            &  90.76 $\pm$ 4.39            \\ & Full supervision-ResUNet \cite{Zhang_2018} &      \textbf{95.26 $\pm$ 4.82}  & \textbf{97.50 $\pm$ 2.84}            &  \textbf{98.38 $\pm$ 2.01}            \\\hline
\end{tabular}
\label{tab:full-seg}
\end{table}

\section{Discussion}

To the best of our knowledge, our proposed MedCLIP-SAM presents the first framework that integrates CLIP and SAM models toward universal radiological segmentation. By leveraging the latest CAM technique, gScoreCAM, which is used in medical imaging for the first time, our method offers a unique solution that allows text-prompt-based interaction, easy adaptation to new data domains/tasks, and data-efficient model (pre-)training. One major contribution of this work lies in the newly devised DHN-NCE loss, which benefits from the synergy of DCL and HN-NCE and has been demonstrated to outperform the state-of-the-art loss functions (see Table \ref{tab:zero acc}) to efficiently fine-tune the BiomedCLIP model with a small batch size. Although we only demonstrated its application in unsupervised CLIP model fine-tuning, we will test its application in full model training in the near future. When using BiomedCLIP and gScoreCAM to obtain saliency maps, we used more simplistic keywords for segmentation tasks, such as ``brain tumor". However, we also noticed that the quality of these maps could benefit from more sophisticated text prompt engineering, including detailed descriptions (e.g., shape and location of the target anatomy/pathology). This leaves an interesting application of our MedCLIP-SAM framework for interactive radiological education. From the ablation studies, both gScoreCAM and fine-tuned BiomedCLIP positively contributed to the success of our method. Our weakly supervised segmentation only improved the accuracy in X-ray-based lung segmentation. This could be explained by the complex contrast of ultrasound and the 3D nature of the brain MRI, which may be more suitable for 3D segmentation. Notably, the latest MedSAM \cite{Ma2023SegmentAI} has demonstrated superior performance for medical applications. However, as it was fine-tuned on large amounts of public medical datasets, which include our test databases, adopting it for our framework will invalidate the ``zero-shot" setting. With encouraging results from SAM in our framework, we aim to further explore the incorporation of MedSAM into MedCLIP-SAM to verify the potential performance enhancement. Finally, we only tested three segmentation tasks and image modalities in this study, and will expand our validation to a broader range of applications and image types.

\vspace{-0.1cm}
\section{Conclusion}
We proposed MedCLIP-SAM, a novel framework that combines CLIP and SAM foundation models to obtain text-prompt-based universal medical image segmentation. The interactive nature of the method provides a unique venue to allow human interaction.  In addition, our newly proposed DHN-NCE loss could potentially benefit broader applications. Our comprehensive experiments demonstrated excellent performance of the proposed framework, which possesses great potential for clinical adoption upon future improvements.

\vspace{-0.1cm}
\subsubsection{\ackname} We acknowledge the support of the Natural Sciences and Engineering Research Council of Canada (NSERC).

\bibliographystyle{splncs04}
\bibliography{bibliography}
%




\end{document}